\documentclass[11pt]{article}
\usepackage[utf8]{inputenc}
\usepackage[dvipsnames]{xcolor}

\usepackage{amsmath}
\usepackage{amssymb}
\usepackage{graphicx}  %
\usepackage{paralist}
\usepackage{multirow}
\usepackage{floatrow}
\usepackage{booktabs}
\usepackage{algorithm}
\usepackage{enumitem}
\usepackage{wrapfig}
\usepackage[caption=false]{subfig}
\usepackage[noend]{algpseudocode}
\usepackage[hidelinks]{hyperref}

\DeclareMathOperator*{\argmax}{arg\,max}
\usepackage{setspace} 

\usepackage{IEEEtrantools}

\newfloatcommand{capbtabbox}{table}[\FBheight][\FBwidth]
\newfloatcommand{capbfigbox}{figure}[\FBheight][\FBwidth]
\usepackage{natbib}
\usepackage[final]{neurips}

\title{Open Set Medical Diagnosis}

\author{
    \textbf{Viraj Prabhu}\thanks{\,\,\,Work done as research intern at Curai.}$\,\,^{,1}$ \qquad
    \textbf{Anitha Kannan$^3$} \qquad
    \textbf{Geoffrey J. Tso$^3$} \qquad 
    \textbf{Namit Katariya$^3$} \qquad \\
    \textbf{Manish Chablani$^3$} \qquad 
    \textbf{David Sontag}\thanks{\,\,\,Work done as advisor to Curai.}$\,\,^{,2}$ \qquad
    \textbf{Xavier Amatriain$^3$} \qquad \\
    $^1$Georgia Tech \qquad
    $^2$MIT \qquad
    $^3$Curai \\
}

\begin{document}

\maketitle

\begin{abstract}
Machine-learned diagnosis models have shown promise as medical aides but are trained under a closed-set assumption, i.e. that models will only encounter conditions on which they have been trained. However, it is practically infeasible to obtain sufficient training data for every human condition, and once deployed such models will invariably face previously unseen conditions. We frame machine-learned diagnosis as an \emph{open-set} learning problem, and study how state-of-the-art approaches compare. Further, we extend our study to a setting where training data is distributed across several healthcare sites that do not allow data pooling, and experiment with different strategies of building open-set diagnostic ensembles. Across both settings, we observe consistent gains from explicitly modeling unseen conditions, but find the optimal training strategy to vary across settings.
\end{abstract}

\section{Introduction}

An increasing number of adults in the US are turning to the internet to find answers to their medical concerns. A survey conducted in \cite{semigran2015evaluation} revealed that in 2012, 35\% of U.S. adults had gone online at least once to self-diagnose. 
In fact, around 7\% of Google’s daily searches are health related \citep{googHealthQueries}. 

To service this need, several online ``symptom checking'' services have emerged, which typically first ask patients a series of questions about their symptoms, and then provide a diagnosis. These services can improve both accessibility as well as provide patients with directed information to guide their medical decision-making~\citep{semigran2015evaluation}. Symptom checkers are increasingly powered by machine-learned diagnosis models. These models are not only showing promise as potential decision aides for patients and medical professionals alike but are also poised to revolutionize patient-facing telehealth services that could move from the current rules-based protocols for nurse hotlines to more accurate and scalable AI systems.

Existing models for clinical decision support (CDS) make a \emph{closed-world assumption} i.e. the universe of diseases is limited to those that have been encoded in the model. In practice, it is likely that a deployed diagnosis model will encounter previously unseen conditions rendering the original assumption unrealistic. Not only is the number of possible diagnoses very large (over 14025 diagnosis codes exist in ICD 9/10~\footnote{Though some codes can be collapsed due to clinical similarity, the actual number is still in a few thousands.}), but obtaining sufficient training data for each condition is also challenging.
As a result, many telehealth providers constrain the coverage of their CDS system to specific areas of care. However, determining whether or not a patient falls within diagnostic scope based on symptoms alone necessitates employing additional models or human expertise, which can be both challenging and expensive. Further, each misdiagnosis is a missed opportunity for better care, and can even be safety-critical in some cases.

Prior work in machine learning has studied the \emph{open-set learning} problem (and the related problem of learning with a reject option), which is concerned with developing approaches that are aware of and can avoid misclassifying previously unseen classes [{\it c.f }~\cite{chow1970optimum, bendale2015towards}].  In this work, we frame diagnosis as an open-set classification problem, and compare the efficacy of different approaches.

Another critical challenge in building diagnosis models is \emph{access to data}. Health data usually lives in hospital repositories and for privacy reasons, can often not be taken outside its respective source site to be pooled with other sources. This makes training models (particularly data-inefficient neural networks) hard. Further, different healthcare sites may have complementary data -- for instance, electronic health records (EHR) of tropical countries are more likely to contain a lot more clinical cases for malaria than the rest of the world. Similarly, hospitals on the US east coast are likely to have more patient encounters for hypothermia than on the west. 
To develop comprehensive and accurate models that cover a wide range of diseases, we need mechanisms to bridge models trained on these individual sites. To this end, we introduce the task of Ensembled Open Set Diagnosis, where we compare methods to ensemble models trained on data sources that cannot be shared, and evaluate their open-set diagnostic performance. While this setup has widespread applicability in healthcare, to the best of our knowledge we are the first to study it.

Our contributions are two-fold:
\begin{enumerate}
    \item We frame machine-learned diagnosis as an open-set learning problem, and study how well existing approaches to open-set learning translate to clinical diagnosis. We find that a simple approach (using an additional ``background'' class) outperforms a state-of-the-art open-set learning. Moreover, approaches that explicitly account for unseen conditions consistently outperform baselines that do not.
    \item We introduce the task of ensembled open-set diagnosis, where the goal is to build an ensemble capable of open-set diagnosis of a target set of conditions, by combining experts trained at different healthcare sites. Each expert contributes a subset of the target conditions, but data cannot be pooled across experts. We find that simple ensembling techniques combined with open-set learning approaches perform well in practice, though we do not find a single winner across all settings.
\end{enumerate}

\section{Setup}

\label{sec:setup}
\begin{figure}
 \centering 
 \includegraphics[width=0.8\textwidth]{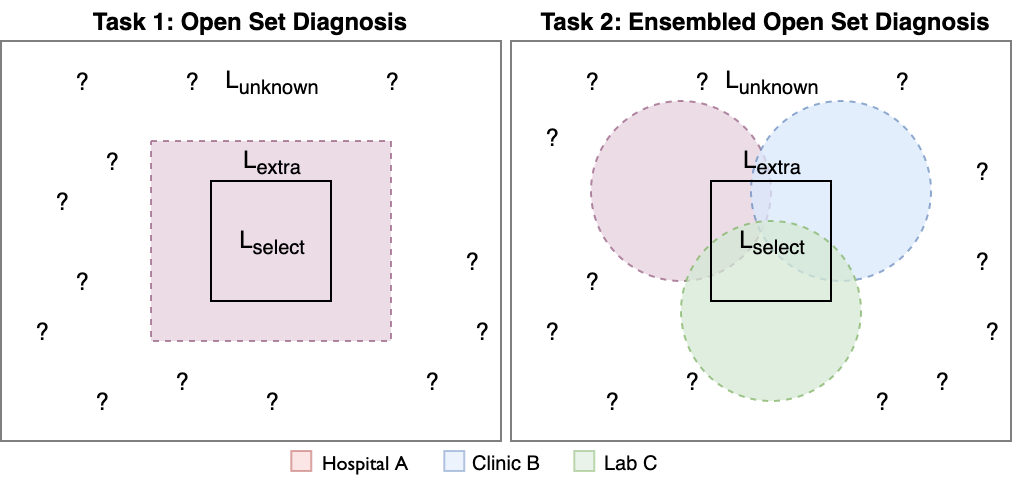}
 \caption{Left: Open Set Diagnosis. The goal is to learn a model to diagnose diseases from $L_{select}$ and reject unseen conditions $\in L_{unknown}$ given a training set of diseases $L_{select}$, and optionally additional data from $L_{extra}$. Right: Ensembled Open Set Diagnosis. In this setting, the goal and evaluation setting is identical to the previous task but training data is now \emph{distributed} across multiple sites. See \S~\ref{sec:setup} for more details.}
\label{fig:teaser}

\end{figure}
Let $\mathcal{Y}$ represent the universe of all possible human diseases that can be diagnosed. 
Let $L_{select} \subset \mathcal{Y}$ represent a subset of diseases for which we have labeled data and wish to include within the scope of diagnosis of our model. For instance, in a telehealth setting, $L_{select}$ could correspond to the subset of diseases that can easily be diagnosed remotely. However, once deployed, this model might encounter cases belonging to any disease $D \in \mathcal{Y}$. Our objective is to develop a model that can correctly diagnose diseases belonging to $L_{select}$, and declare \textrm{NOTA} (none of the above) otherwise. 

Let $\mathcal{U}= \mathcal{ Y} \setminus L_{select} $ be the set of all possible conditions that our model needs to reject, {\it i.e.} declare \textrm{NOTA}. We further divide $\mathcal{U} = L_{unknown} \cup L_{extra}$ where $L_{unknown} \cap L_{extra} = \emptyset$, with $L_{extra}$ representing some additional ``extra'' conditions that are outside the scope of diagnosis, but for which we may have some small amount of training data. We can view $L_{extra}$ as a proxy for unseen classes during training. 

Note, importantly, that cases corresponding to $L_{unknown}$ are only seen at test time. In a telehealth medical setting, $L_{unknown}$ may correspond to the set of conditions that are rare or challenging to obtain training data for, but that the model might potentially encounter once deployed. In this \emph{open-set} setting, we want to prevent misdiagnosis and instead recommend additional diagnostic evaluation such as a physical examination, laboratory tests, or imaging studies.

We study two experimental tasks within this setting:

\par \noindent
 \textbf{Task 1: Open Set Diagnosis.} In this task (see Fig~\ref{fig:teaser}, left), we assume centralized access to training data, and attempt to learn a medical diagnosis model that can accurately diagnose a given clinical case as either one of $L_{select}$, or as \texttt{NOTA} (none of the above).
 
\par \noindent
\textbf{Task 2: Ensembled Open Set Diagnosis.}
In this task (see Fig~\ref{fig:teaser}, right), we assume that training data is ~\emph{distributed} across $K$ sites (say a hospital, a clinic, and a specialty lab) that do not allow data-pooling. Each site is provided with clinical case data spanning a label set $L_i$ and trains a corresponding expert model $M_{i}$. Each $L_i$ comprises of a relevant subset $L_{i_{rel}} \subset L_{select}$, and optionally an `extra' subset $L_{i_{extra}} \cap\  L_{select} = \phi$. As shown in Fig.~\ref{fig:teaser}, these label spaces may have overlap. The goal is to train an \emph{ensemble model} of these individual experts that is capable of diagnosing $L_{select}$ clinical conditions, while avoiding misdiagnosis of conditions in $L_{unknown}$.

\section{Approach}
We experiment with three approaches to open-set classification -- ``vanilla'' softmax cross-entropy with thresholding, training an explicit ``background'' class, and a recently proposed state-of-the-art approach based on neural network ``agnostophobia''~\citep{dhamija2018reducing}. 

\subsection{Models for Open Set classification}
\label{sec:loss}
\par \noindent
\textbf{Cross-entropy (\texttt{CE}) loss with confidence thresholding}: In this approach ({\it c.f.} \cite{Matan90, Giorgio2002}), a classification model $f: \mathcal{X} \mapsto \mathcal{L}_{select}$ is learned.
Assuming that the model will have high predictive softmax entropy for datapoints belonging to previously unseen classes, all datapoints for which the model's confidence falls below a threshold $\theta$ (picked using a validation set) are classified as \textrm{NOTA}. In particular, for input {\bf x}, with predicted probability distribution, $P(c|{\bf x})$, a prediction $c^*$ is made as follows:
\begin{equation}
    c^* = \begin{cases} \argmax_c P(c|{\bf x}), & \text{if}\ P(c|{\bf x}) \ge \theta \\ \textrm{NOTA}  \end{cases}, 
\label{eq:ce}
\end{equation}

\par \noindent
 \textbf{Background (\texttt{BG}) class}:
In this line of work, an explicit `background' class is included as a catch-all class for modeling out-of-domain inputs. An assumption here is the availability of a set of examples that are sufficiently representative of unknown classes.
In our setup, this would mean training a classifier $f: \mathcal{X} \mapsto \mathcal{L}_{select}+1$ where the \textrm{NOTA} i.e. background class is trained using data from $L_{extra}$ as a proxy for unseen classes.
 
\par \noindent
\textbf{Entropic Open-set (\texttt{EOS}) Loss}: Introduced in ~\cite{dhamija2018reducing}, this loss encourages high predictive entropy for examples corresponding to unseen classes. Similar to the previous approach, $L_{extra}$ is used as a proxy for unseen conditions at training time. Thus, a classifier  $f: \mathcal{X} \mapsto \mathcal{L}_{select}$ is learned with the entropic loss $J_{eos}$ defined for datapoint $(x, c)$ as follows:
\begin{equation}
    J_{eos} = \begin{cases} -\log P(c|{\bf x}),  & \text{if}\ x \in L_{select} \\ -\frac{1}{C} \Sigma_{c=1}^{|L_{select}|} \log P(c|{\bf x}), & \text{if}\ x \in L_{extra} \end{cases}
\end{equation}
The intuition is to encourage high predictive entropy on unseen (i.e. $L_{extra}$) examples, and train with regular cross-entropy on seen examples. Similar to \texttt{CE}, prediction follows Eq.~\ref{eq:ce} and \textrm{NOTA} is predicted via confidence thresholding. We also tried the Objectosphere loss~\citep{dhamija2018reducing} that builds on \texttt{EOS} by additionally encouraging a margin between feature activation magnitudes for knowns and unknowns, but did not observe performance gains. In our experiments we present results using the \texttt{EOS} loss.

Another applicable line of work compared to in~\cite{dhamija2018reducing} are approaches that explicitly model network uncertainty~\citep{gal2016dropout,lakshminarayanan2017simple}. However, they find the \texttt{CE}, \texttt{BG}, and \texttt{EOS} approaches to significantly outperform such a baseline (specifically, \emph{deep ensembles} proposed in~\cite{lakshminarayanan2017simple}) on the open set classification task. As a result, we do not study these in our paper. Further, as the same work points out, uncertainty estimation is an orthogonal approach that can potentially be combined with the approaches we study.

\subsection{Ensembling strategies}
To ensemble diagnosis models trained on different sites (Task 2), we experiment with two strategies:
\par \noindent
{\bf Max-confidence (\texttt{naive})}:  
We predict the class with highest confidence as the ensemble prediction. For experts trained with \texttt{BG}, we average confidence for the background class across experts.

\par \noindent
{\bf Mixture of Experts (\texttt{learned})}: We assume access to a very small set of pooled and previously heldout training data for training an ensemble. This dataset can come from each site sharing a small amount of de-identified data, or through manual curation of clinical cases. In this setup, we parameterize our ensemble with a ``mixture of experts''~\citep{jacobs1991adaptive} architecture that makes use of a fully connected (FC) layer as a gating function over the current input, that is elementwise multiplied with a concatenation of logits from each expert for the same input, and passed through another FC layer.
We experiment with \texttt{CE}, \texttt{BG}, and \texttt{EOS} losses for training this \texttt{learned} ensemble.

\section{Experimental Setup}
\subsection{Clinical case simulation}
\label{sec:simulation}
We simulate a large number of clinical vignettes from a medical decision expert system \citep{miller_internist-1_1982} to use as our dataset. The expert system has a knowledge base of diseases, findings (covering symptoms, signs, and demographic variables), and their relationships. Relationships between finding-disease pairs are encoded as \emph{evoking strength} and \emph{frequency}, with the former indicating the strength of association between the constituent finding-disease pair and the latter representing frequency of the symptom in patients with the given disease. Further, \emph{disease prevalence} metadata suggests whether a given disease is very common, common, or rare. 

The simulation algorithm \citep{qmr-simulated-cases, ravuri18} makes a closed world assumption with the universe of diseases (denoted $\mathcal{Y}$) and findings ($\mathcal{F}$ ) being those in the knowledge base.  The simulator first samples a disease $d \in \mathcal{Y}$ and demographic variables, and then samples findings in proportion to frequency for the picked disease. Each sampled finding is assigned to be present or absent, based on frequency. If assigned present, then findings that are impossible to co-occur are removed from consideration (\emph{e.g.} a person cannot have both productive and dry cough). The simulation for a case ends when all findings in the knowledge base have been considered. At the end of the simulation, a clinical case is a pair ({\bf x}, d) where  $d \in \mathcal{Y} $ is the diagnosis and {\bf x} captures the instantiated finding. In particular, each element $x_{j}$ is a binary variable of finding presence. For our experiments, we limit to demographic variables and symptoms as these are the most likely available findings when first diagnosing a patient in a telehealth setting; we also restrict cases to 5-8 symptoms reflecting a typical clinical case.

\subsection{Dataset construction}
\label{sec:dataset}

\par \noindent
{\bf Constructing $L_{select}$, $L_{unknown}$ and $L_{extra}$:} As most telehealth services are likely to include common conditions within diagnostic scope, we choose $L_{select}$ to be the 160 diseases marked as ``very common'' in the knowledge base of the clinical case simulator described above. Further, recall that we want to be robust to misdiagnosis of previously unseen and possibly rare conditions, including (and especially) those with high symptom overlap with seen $L_{select}$. Therefore, we construct a challenging $L_{unknown}$ split as follows: First, we average one-hot encodings (using a $D=2052$ dimensional vocabulary) of the cases for all (=830) diseases in our knowledge base. Then, we apply dimensionality reduction via principal component analysis on this $N \times D$ dimensional matrix, retaining $D'=500$ components that explain 90\% of total variance. We pick the first unique nearest neighbor to each condition in $L_{select}$ to yield 160 unseen conditions that have high finding overlap with $L_{select}$.  This constitutes our $L_{unknown}$ set of diseases. Finally, 160 diseases corresponding to $L_{extra}$ are chosen uniformly at random from the set of conditions that remain. 

An example of a challenging (disease, distractor) pair between $L_{select}$ and $L_{unknown}$ that we obtain via the scheme described is `Amblyopia' (lazy eye) and `Diabetic Ophthalmoplegia' -- both conditions are vision impairments (often, double vision problems) that lead to blurred vision. Another pair is `Actinic Keratosis' and `Melanoma', where the former is a scaly patch on the skin due to prolonged sun exposure, while melanoma manifests as an unusual skin growth. Yet another example is (`Melancholia', `Bipolar disorder') both of which share symptoms such as depressed moods and anxiety, except that bipolar disorder tends to be episodic and requires longitudinal insight.

\par \noindent {\bf Data split for Task 1.} We simulate clinical cases employing the strategy described in \S.~\ref{sec:simulation}. 
For diseases in $L_{select}$, we simulate 1000 cases per condition. From this, we use 20\% for testing ($D_{select}$), and the remaining 80\% ($D'_{select}$)  for training and validation. To mimic the difficulty of obtaining training data for less common conditions, we only simulate 100 cases on average for each condition in $L_{extra}$, and use it for training \texttt{BG} and \texttt{EOS} models. Finally, we simulate 1000 cases for each condition in $L_{unknown}$ to obtain $D_{unknown}$. We report performance over our test set constructed as $D_{select} \bigcup D_{unknown}$.

\par \noindent {\bf  Data split for Task 2.} For this setting, we simulate a realistic distribution of data among $M=4$ individual healthcare sites. As previously discussed, each site contributes cases spanning a subset of relevant diseases $L_{i_{rel}} \subset L_{select}$. We achieve this by dividing $L_{select}$ (from Task 1) across the $M$ sites, and vary the degree of overlap between different sites. We define overlap as the number of conditions that occur in $> 1$ sites, as a \% of $|L_{select}|$. Further, each site also contributes ``extra'' diseases $L_{i_{extra}} \not\subset L_{select}$ that are out of the target diagnostic scope. We pick conditions uniformly at random from the set of remaining conditions as $L_{i_{extra}}$. %

\subsection{Metrics}
For open-set diagnosis, we use the Open-Set Classification Rate (OSCR) metric proposed in~\cite{dhamija2018reducing}, which plots false positive rate (FPR) versus correct classification rate (CCR) as a function of confidence threshold $\theta$:
\small
$$\operatorname{FPR}(\theta)=\frac{\left|\left\{x\left|x \in \mathcal{D}_{unknown} \wedge \max _{c} P(c | x) > \theta\right\} |\right.\right.}{\left|\mathcal{D}_{unknown}\right|}$$
$$\operatorname{CCR}(\theta)=\frac{\left|\left\{x\left|x \in \mathcal{D}_{select} \wedge \arg \max _{c} P(c | x)=\hat{c} \wedge P(\hat{c} | x)>\theta\right\} |\right.\right.}{\left|\mathcal{D}_{select}\right|}$$
\normalsize
FPR measures the fraction of unknown examples that are misclassified as one of the $L_{select}$ classes; a high FPR indicates that unknown diseases are often conflated with a known class. Meanwhile, CCR directly measures the fraction of known examples that are classified correctly. Ideally, a robust and accurate classifier achieves high CCR at low false positive rates. In practice, we compare approaches based on their CCR corresponding to a target FPR. The OSCR metric overcomes many of the shortcomings of previously proposed open-set metrics, such as accuracy vs confidence, AUROC, and recall@K. We refer readers to Section 4.3 in ~\cite{dhamija2018reducing} for a detailed comparison.

\subsection{Base Model}

For Task 1, we parameterize models as 2-layer Multilayer Perceptrons (MLP's) with one-hot feature encodings, using a global vocabulary constructed from the union of all case findings. We use ReLU non-linearities and 100 hidden units. As in~\cite{dhamija2018reducing}, we set bias terms in the logit i.e. second hidden layer to 0. For ensemble models (Task 2), we employ the same architecture for individual site models (experts), but assume access to a single shared vocabulary of symptoms; however, symptoms may be missing across different sites because of the data available to that site.

\section{Results}
All models are trained with early stopping based on validation loss. We use an initial learning rate of $10^{-3}$ and use Adam optimization~\citep{kingma2014adam}. Further, since $L_{extra}$ is sampled randomly, and to ensure statistical significance, we train models over three random samplings of $L_{extra}$ and report performance means and standard deviations.

\subsection{Task 1: Open Set diagnosis}
\label{sec:results}
To study the difference between different algorithms, we choose an operating threshold corresponding to FPR values of 10-30\%, and report the corresponding CCR. This choice allows capturing an important trade-off; in our setting, we assume that it is better to misclassify a known ``common'' disease as unknown (\textrm{NOTA}) than to misdiagnose an unknown ``rare'' disease as known (one of $L_{select}$). Further, we note that while our choice of operating false positive rates appears large, they are commensurate with our extremely challenging and large ($>204k$ examples) test set. 

\begin{figure}
\begin{floatrow}
\floatbox{table}[.6\textwidth][\FBheight][t]
{\vspace{15pt}\caption{Open Set Diagnostic performance. Error bars denote standard deviation over 3 random samplings of $L_{extra}$.}\label{tab:osd}}
{\resizebox{0.6\textwidth}{!}{
\begin{tabular}{lccc}
\toprule
 &  & $\uparrow$ CCR@FPR of   & \\
Algorithm & 0.1 & 0.2   & 0.3   \\ \hline
\texttt{CE} & 74.81 \scriptsize{$\pm$ 0.23} & 89.50 \scriptsize{$\pm$ 0.05} & 93.40 \scriptsize{$\pm$ 0.13} \\
\texttt{BG} & \textbf{79.25 \scriptsize{$\pm$ 0.22}} & \textbf{92.30 \scriptsize{$\pm$ 0.12}} & \textbf{95.75 \scriptsize{$\pm$ 0.05}} \\
\texttt{EOS} & 75.21 \scriptsize{$\pm$ 1.01} & 90.75 \scriptsize{$\pm$ 0.12} & 94.70 \scriptsize{$\pm$ 0.13} \\
\bottomrule
\end{tabular}}}
\floatbox{figure}[.3\textwidth][\FBheight][t]
{\includegraphics[width=0.33\textwidth]{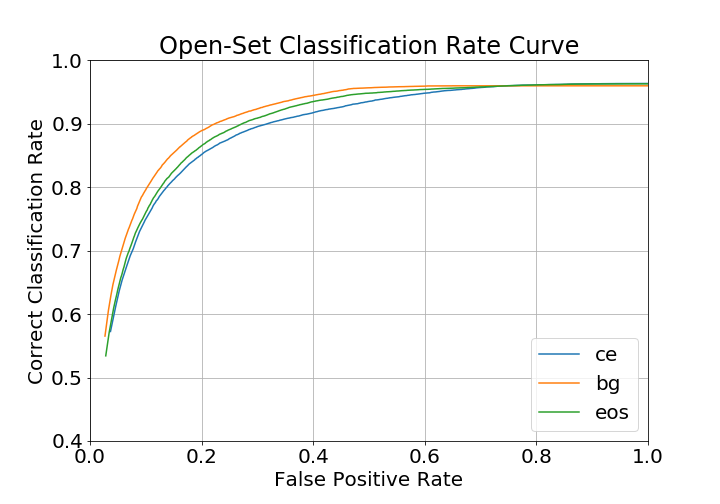}}
{\caption{Open Set Diagnosis OSCR curve.}\label{fig:osd_oscr}}
\end{floatrow}
\end{figure}

Table~\ref{tab:osd} compares different methods at three different FPR values. We can see that methods that explicitly model unseen classes perform better than methods that do not. Interestingly, and in contrast to the findings in \cite{dhamija2018reducing}, we find \texttt{BG} to outperform \texttt{EOS} in this setting~\footnote{To be clear, this is not an apples-to-apples comparison as \cite{dhamija2018reducing} study a considerably different task and setup.}. The OSCR curve in Figure~\ref{fig:osd_oscr} corroborates this trend across thresholds. These results suggest that, consistent with prior work, explicitly modeling out-of-distribution conditions is beneficial when the test time evaluation is open set, though the optimal choice of modeling strategy may be task dependent.

\small
\normalsize

\subsection{Task 2: Ensembled Open Set Diagnosis}

Table~\ref{tab:open} compares the performance corresponding to the two methods for ensembling individual experts (with different loss functions) trained at various sites. Additionally, as a performance upper bound, for each setting we also mark the performance of an ``oracle'' \texttt{BG} model that has \emph{centralized} access to all the training data.

\begin{table}[]
\resizebox{0.8\textwidth}{!}{
\begin{tabular}{llccc}
\toprule
        &  &  & $\uparrow$ CCR@FPR of & \\    
          & Algorithm & 0.1 & 0.2 & 0.3 \\ \hline
          & \texttt{CE} & \--- & 86.10 \scriptsize{$\pm$ 0.27} & 89.64 \scriptsize{$\pm$ 0.23}     \\
\texttt{naive} & \texttt{BG} & \textbf{74.25 \scriptsize{$\pm$ 0.50}} & \textbf{88.38 \scriptsize{$\pm$ 0.19}} & \textbf{91.07 \scriptsize{$\pm$ 0.12}}  \\
          & \texttt{EOS} &  67.49 \scriptsize{$\pm$ 1.42} & 84.44 \scriptsize{$\pm$ 0.77} & 88.30 \scriptsize{$\pm$ 0.55}  \\ 
          & \textcolor{orange}{\texttt{oracle}} & 79.25 \scriptsize{$\pm$ 0.22} & 92.30 \scriptsize{$\pm$ 0.12}   & 95.75 \scriptsize{$\pm$ 0.05}  \\ 
          \hline
          & \texttt{CE+CE}           &  73.90 \scriptsize{$\pm$ 1.10} & 87.93 \scriptsize{$\pm$ 0.45} & 92.16 \scriptsize{$\pm$ 0.38}  \\
          & \texttt{BG+CE}          & 72.83 \scriptsize{$\pm$ 0.80} & 87.20 \scriptsize{$\pm$ 0.49} & 91.49 \scriptsize{$\pm$ 0.48}    \\
\texttt{learned} & \texttt{EOS+CE}           &   \textbf{76.23 \scriptsize{$\pm$ 0.52}} & 89.00 \scriptsize{$\pm$ 0.33} & \textbf{92.67 \scriptsize{$\pm$ 0.27}}   \\
          & \texttt{EOS+BG}          & \textbf{77.26 \scriptsize{$\pm$ 0.91}} & \textbf{89.90 \scriptsize{$\pm$ 0.34}} & \textbf{92.10 \scriptsize{$\pm$ 0.66}}  \\
          & \texttt{EOS+EOS}          &  74.51 \scriptsize{$\pm$ 0.18} & 88.31 \scriptsize{$\pm$ 0.04} & 91.51 \scriptsize{$\pm$ 0.13}   \\
          & \textcolor{orange}{\texttt{oracle}}          &  79.75 \scriptsize{$\pm$ 0.18} & 92.54 \scriptsize{$\pm$ 0.24} & 95.79 \scriptsize{$\pm$ 0.12}    \\\hline
\end{tabular}}
\caption{Ensembled Open Set Diagnosis performance for \texttt{naive} and \texttt{learned} approaches. Error bars denote standard deviation over 3 random samplings of $L_{extra}$.}\label{tab:open}
\end{table}

We observe that across all approaches, models that explicitly model unseen conditions (\texttt{BG} and \texttt{EOS}) consistently outperform those that do not (\texttt{CE}). Further, we notice opposing trends across the \texttt{naive} and \texttt{learned} ensembles -- in the former, \texttt{BG} significantly outperforms \texttt{EOS} (row 2 vs 3), while in the latter we observe the converse (row 5 vs 6). 

Training the \texttt{learned} ensemble with the \texttt{EOS} loss on top of experts trained with \texttt{BG} appears to improve mean performance, but doing so with the \texttt{EOS} loss does not (rows 7-9). Finally, in both \texttt{naive} and \texttt{learned} cases, the \texttt{oracle} significantly outperforms all approaches, with the gap representing the error introduced by the distributed training and ensembling.

\begin{figure*}[th!]
    \centering
    \begin{minipage}[b]{.45\linewidth}
        \centering
        \includegraphics[width=\linewidth]{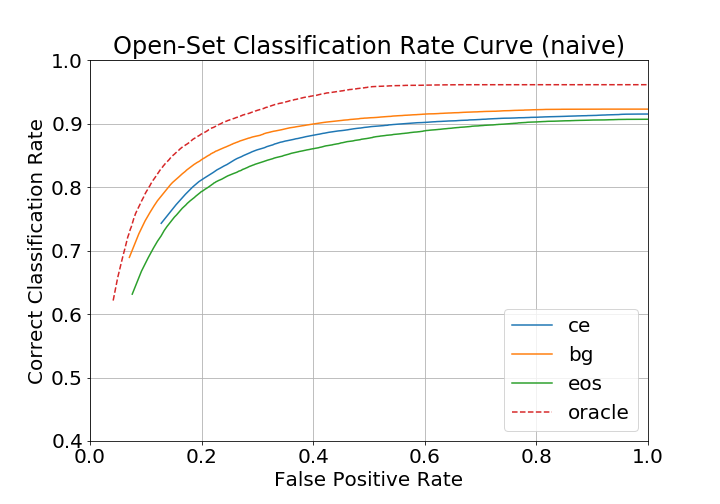}
        (a)
    \end{minipage}\hfill
    \begin{minipage}[b]{.45\linewidth}
        \centering
        \includegraphics[width=\linewidth]{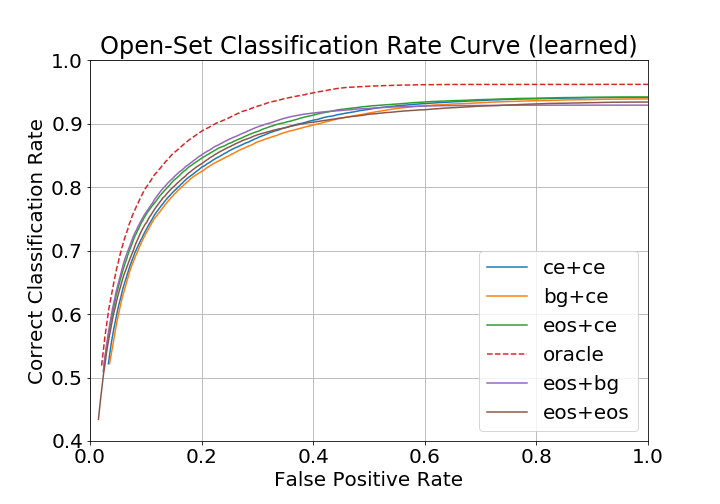}
        (b)
    \end{minipage}\hfill
\caption{Ensembled Open Set Diagnosis: OSCR curves for (a) \texttt{naive} and (b) \texttt{learned} approaches.}
    \label{fig:oscr_learned}
    \vspace{-5pt}
\end{figure*}

In Fig.~\ref{fig:oscr_learned}, we present OSCR curves for our studied approaches. Clearly, different models have different starting operating points for FPR. We further break down errors for known examples (i.e. CCR performance) into misclassification as background vs as an incorrect foreground class. For instance, we find the 24.01\% CCR error@FPR=0.1 for (1 of 3 runs of) our \texttt{learned} \texttt{EOS+BG} model breaks down as 23.54\% and 0.47\%, respectively. We find similar trends to hold across approaches. Clearly, the model is very accurate at distinguishing between classes it has been trained on, but still struggles with consistently rejecting previously unseen conditions.

\begin{figure}
\begin{floatrow}
\floatbox{table}[.55\textwidth][\FBheight][t]
{\caption{Ensembled Open Set Diagnosis performance (CCR@FPR=0.1) across varying inter-expert overlap.}\label{tab:overlap}}
{\resizebox{0.55\textwidth}{!}{
\begin{tabular}{llccc}
\toprule
        &  &  & \% overlap & \\    
          & Algorithm & 0\% & 50\% & 100\% \\ \hline
          & \texttt{CE} & \--- & \--- & \---     \\
\texttt{naive} & \texttt{BG} & \textbf{70.61 $\pm$ 1.62} & \textbf{74.25 $\pm$ 0.50} & \textbf{74.40 $\pm$ 0.59}  \\
          & \texttt{EOS} &  63.83 $\pm$ 0.29 & 67.49 $\pm$ 1.42 & 68.74 $\pm$ 1.45  \\ 
          \hline
          & \texttt{CE+CE}           & 73.20 $\pm$ 0.80 & 73.90 $\pm$ 1.10 & 73.76 $\pm$ 0.29  \\
          & \texttt{BG+CE}          & 71.17 $\pm$ 0.41 & 72.83 $\pm$ 0.80 & 72.15 $\pm$ 1.10     \\
\texttt{learned} & \texttt{EOS+CE}           & \textbf{75.07 $\pm$ 1.04}   & \textbf{76.23 $\pm$ 0.52} &  75.58 $\pm$ 0.81  \\
          & \texttt{EOS+BG}          & \textbf{75.69 $\pm$ 1.08} & \textbf{77.26 $\pm$ 0.91} & \textbf{77.03 $\pm$ 0.67}   \\
          & \texttt{EOS+EOS}          & 72.94 $\pm$ 0.34  & 74.51 $\pm$ 0.18 & 73.84 $\pm$ 0.76   \\
        \hline
\end{tabular}}}
\floatbox{figure}[.45\textwidth][\FBheight][t]
{\includegraphics[width=\linewidth]{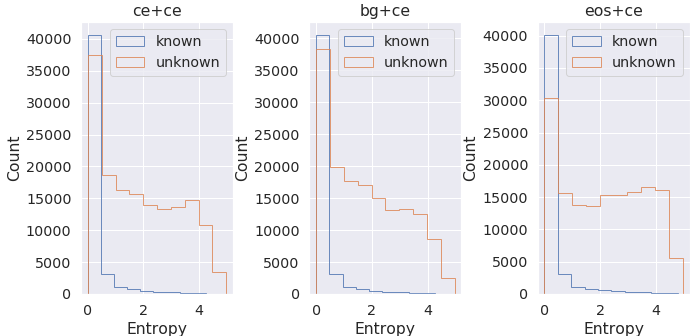}}
{\label{fig:entropy}\caption{Histograms of softmax entropy across models over our test set.}}
\end{floatrow}
\end{figure}

In Fig.~\ref{fig:entropy}, we plot the histograms of softmax \emph{entropy} over our test set from the \texttt{learned} \texttt{CE+CE}, \texttt{BG+CE}, and \texttt{EOS+CE} models. As expected, we find that models trained with \texttt{EOS} losses have higher predictive entropy for unseen examples. Note that we do not observe a completely clear separation even with the \texttt{EOS} loss, which we attribute to the difficulty of our test set.





\small
\normalsize

\textbf{Varying overlap.}
We study performance at $0\%$, $50\%$, and $100\%$ overlap (number of conditions that occur in $> 1$ sites, as a \% of $|L_{select}|$). Correspondingly, the number of conditions per expert ranges from $|L_{select}|/M=40$ conditions ($0\%$ overlap) to $80$ (100\% overlap). As seen in Table~\ref{tab:overlap}, we observe near-consistent trends across all degrees of overlap.

\begin{table}[]
\centering
\resizebox{1\textwidth}{!}{
\setlength\tabcolsep{2pt}\begin{tabular}{llllll}
\toprule
Findings                                                                                                                                                            & Label                                                            & \texttt{CE+CE}                                                                                                             & \texttt{BG+CE}                                                                                                              & \texttt{EOS+CE}                                                                                                               & \texttt{EOS+BG}                                                                                                           \\ \hline
\begin{tabular}[c]{@{}l@{}}middle age (40-70 years)\\ female\\ sudden onset of symptoms\\ muscle rigidity\\ trauma\\ muscle spasm\\ jaw pain\end{tabular}           & \begin{tabular}[c]{@{}l@{}}NOTA \\(tetanus)\end{tabular}                                                   & \begin{tabular}[c]{@{}l@{}}\color{Red}{whiplash injury (95.28)}\\ scarlet fever (2.71)\\ diverticulitis (0.39)\end{tabular}    & \color{ForestGreen}
{NOTA}                                                                                                               & \color{ForestGreen}{NOTA}                                                                                                                 & \color{ForestGreen}{NOTA}                                                                                                             \\ \hline
\begin{tabular}[c]{@{}l@{}}newborn (\textless{}2 months)\\ male\\ adrenal insufficiency\\ weakness, generalized\\ anorexia\\ skin pigmentation changes\end{tabular} & \begin{tabular}[c]{@{}l@{}}NOTA\\ (addison disease)\end{tabular} & \begin{tabular}[c]{@{}l@{}}\color{Red}{viral pneumonia (99.39)}\\ acute tonsilititis (0.08)\\ appendicitis (0.08)\end{tabular} & \begin{tabular}[c]{@{}l@{}}\color{Red}{viral pneumonia (99.67)}\\ cholecystitis (0.06)\\ acute tonsillitis (0.03)\end{tabular} & \color{ForestGreen}{NOTA}                                                                                                                 & \color{ForestGreen}{NOTA}                                                                                                             \\ \hline
\begin{tabular}[c]{@{}l@{}}adolescent (12 - 18 yrs) \\ female \\ hair loss, patchy \\ sudden onset of symptoms \\ dermatitis atopic \end{tabular}       & \begin{tabular}[c]{@{}l@{}}NOTA \\ (alopecia areata) \end{tabular}                                                             & \begin{tabular}[c]{@{}l@{}} \color{Red}{tinea capitis (93.08)} \\ lateral ankle sprain (2.40) \\ plantar war (0.59) \end{tabular}                                                                                              & \begin{tabular}[c]{@{}l@{}} \color{Red}{tinea capitis (96.22)} \\ lateral ankle sprain (0.93) \\ atopic dermatitis (0.57)\end{tabular}                                                           & \color{ForestGreen}{NOTA}                                                                                                       & \begin{tabular}[c]{@{}l@{}} \color{Red}{tinea capitis (95.29)} \\ chronic urethritis (2.03) \\ atopic dermatitis (0.56) \end{tabular}                                                                                                  \\ \hline
\begin{tabular}[c]{@{}l@{}} child (1-11 years) \\ male \\ localized rash \\ contact w/ similar symptoms \\ few days (2-7) \end{tabular}                                                                                                                               & impetigo                                                                  & \color{red}{NOTA}                                                                                                             & \color{red}{NOTA}                                                                                                                             & \color{red}{NOTA}                                                                                                                               &   \begin{tabular}[c]{@{}l@{}} \color{ForestGreen}{impetigo (79.20)} \\ varicella (20.25) \\ folliculitis (0.23) \end{tabular}                                                                                                                    \\ \hline
                                                                                                                        \begin{tabular}[c]{@{}l@{}}adolescent (~12-18 years) \\ male \\ history of atrial flutter \\ supraventricular tachycardia \\ generalized weakness \end{tabular} & atrial flutter   & \begin{tabular}[c]{@{}l@{}}\color{red}{atrial fibrillation (99.97)} \\  marijuana intoxication (0.01)  \\ paroxysmal supraventricular \\ tachycardia (0.01) \end{tabular} & \begin{tabular}[c]{@{}l@{}}\color{red}{atrial fibrillation (99.98)} \\  marijuana intoxication (0.004)  \\ paroxysmal supraventricular \\ tachycardia (0.004) \end{tabular}  & \begin{tabular}[c]{@{}l@{}}\color{red}{atrial fibrillation (99.72)} \\  marijuana intoxication (0.14)  \\ viral pneumonia (0.04) \end{tabular}  & \begin{tabular}[c]{@{}l@{}}\color{red}{atrial fibrillation (99.70)} \\ viral  pneumonia (0.11)  \\ marijuana intoxication (0.07) \end{tabular}  \\ \bottomrule
\end{tabular}}
\caption{Sample model predictions. Columns 1-2 represent the case findings and ground truth condition, while columns 3-6 show predictions across models, either as top-3 predictions (and corresponding scores), or \textrm{NOTA}. We color code correct predictions in \textcolor{ForestGreen}{green} and incorrect ones in \textcolor{red}{red}.}
\label{fig:qual}
\end{table}

\textbf{Qualitative examples}. Fig.~\ref{fig:qual} qualitatively compares methods.
Row 4 presents a clinical vignette for `alopecia areata' belonging to $L_{unknown}$. While the clinical presentation of this patient has both atopic dermatitis and patchy hair loss, most models appear to ignore patchy hair loss (main indicator of `alopecia areata'), and focus on the sudden onset of `atopic dermatitis'.
Similarly, the clinical vignette corresponding to `atrial flutter' (also in $L_{unknown}$) is misdiagnosed as `atrial fibrillation'.  In fact, atrial fibrillation and atrial flutter tend to often co-occur~\citep{george00}, and so patient symptoms may not be sufficient to differentiate the two. This also sheds light on the complexity of medical diagnosis when multiple diseases share symptoms and also co-manifest in a patient.

{\bf Closed-set diagnosis.} 
Lastly, we measure \emph{closed-set} diagnostic performance i.e. evaluate on a test set consisting of heldout examples from $L_{select}$ alone (=31,000 examples). We use recall@k as our metric. We find that for both \texttt{naive} and \texttt{learned} settings, all approaches perform similarly. We also find this to hold true across degrees of overlap. For example, \texttt{naive} \texttt{CE}, \texttt{BG}, and \texttt{EOS} achieve \{92.05, 99.1\}, \{92.37, 99.23\}, and \{90.91, 99.18\} recall@\{1, 3\} respectively at 50\% overlap. We conclude that explicitly modeling unseen conditions does not adversely affect closed-set performance.

\vspace{-5pt}
\section{Related Work}
\vspace{-5pt}

\textbf{Machine-learning for diagnosis.}  A number of works have proposed machine-learned diagnostic models~\citep{Wang_LR_RiskPrediction, miotto_deep_2016, Ling_DeepRL_Diagnostic, Shickel_Deep_EHR, Rajkomar18, liang2019, ravuri18}. ~\cite{Rajkomar18} propose a deep neural architecture to predict ICD codes from both structured and clinical notes in EHR, while \cite{liang2019} introduce a model for predicting ICD codes for pediatric diseases.~\cite{ravuri18} present an approach to combine knowledge from a clinical medical expert system with electronic health records to learn models for diagnosis. 
Across prior work, the space of diseases considered is fixed across train and test. 

\textbf{Learning with reject option}. The problem of learning with an additional reject option has a long history in the literature~\citep{chow1970optimum,herbei2006classification,bartlett2008classification}, and recent work has extended this to deep networks under various frameworks. ~\cite{bendale2015towards} frame the problem as one of \emph{open-set learning}, and propose an additional OpenMax layer that explicitly estimates the probability of a datapoint being from an unseen class. Others have studied this problem as one of detecting (and rejecting) \emph{out-of-distribution} test datapoints (or outlier rejection). Approaches have included widening separation between in- and out-of-distribution examples via temperature scaling~\citep{liang2017enhancing}, or by encouraging uniform output distributions for unseen examples~\citep{lee2017training,dhamija2018reducing}. Other work has looked at estimating \emph{predictive uncertainty} from deep networks, either by averaging Monte Carlo samples of examples passed through a dropout network ~\citep{gal2016dropout}, or by training ensembles~\citep{lakshminarayanan2017simple}.

\textbf{Ensemble and Federated Learning}. Learning ensembles of models for improving performance and robustness has a long tradition in machine learning [c.f. ~\cite{dietterich2000ensemble}]. Popular strategies have included learning adaptive mixtures of experts~\citep{jacobs1991adaptive}, and adaptive boosting algorithms~\citep{collins2002logistic}. More recently, a related area of interest has been \emph{federated} learning of models from siloed data providers~\citep{li2019federated} in a distributed and privacy-preserving manner, and recent work~\citep{liu2018fadl} has studied this in the context of a healthcare application.

In this work, we combine these three threads, and explore methods to learn diagnosis models with reject options under both centralized and federated settings.

\vspace{-5pt}
\section{Conclusion}
\vspace{-5pt}
In this work we study machine-learned diagnosis as an open-set learning problem where the model must additionally learn to not diagnose when faced with a previously unseen condition. We apply modern methods to this problem in two settings -- first with centralized training data and the second with distributed data across sites that do not permit data-pooling. Across settings, we observe gains from modeling unknown conditions, but find different strategies to be optimal in different settings.

Our work has certain limitations that we will seek to overcome in future work. Firstly, we observe that softmax scores from our diagnostic models tend to be highly peaky, and models will likely benefit from calibration. Further, we only assume identical model families and architectures across sites, whereas in practice we would like to relax this assumption, and potentially ensemble rule-based diagnostic engines with learned systems. In this work we restrict our evaluation to simulated data, and a natural extension would be to benchmark our approach on real-world clinical case data. Finally, we do not model distribution shift, both across sites, and between experts and deployment, both of which are essential challenges to overcome on the road to building reliable diagnostic models.

\small{
\bibliography{main}
}
\end{document}